\documentclass[11pt]{article}

\usepackage[utf8]{inputenc}
\usepackage[T1]{fontenc}
\usepackage{fullpage}
\usepackage{times}
\usepackage{graphicx}
\usepackage{amsmath, amssymb}
\usepackage{booktabs}
\usepackage{microtype}

\PassOptionsToPackage{numbers,sort&compress}{natbib}
\usepackage{natbib}
\usepackage[hidelinks]{hyperref}
\hypersetup{
  pdfauthor={Dongjin Park and Hasung Yeo and JoonWoo Lee},
  pdftitle={OvA-LP: A Simple and Efficient Framework for Federated Learning on Non-IID Data}
}

\title{OvA-LP: A Simple and Efficient Framework for Federated Learning on Non-IID Data}

\author{Dongjin Park, Hasung Yeo, Joon-Woo Lee\thanks{Corresponding Author}\\ Department of Computer Science and Engineering\\ Chung-Ang University\\ \texttt{\{thrudgelmir,sam1705,jwlee2815\}@cau.ac.kr} \\ }
\date{}

\begin{document}
\maketitle

\begin{abstract}
Federated fine-tuning (FFT) adapts foundation models to decentralized data but remains fragile under heterogeneous client distributions due to local drift, i.e., client-level update divergences that induce systematic bias and amplified variance in the global model. Existing aggregation and personalization methods largely correct drift post hoc, which proves brittle under extreme non-IID conditions. We introduce OvA-LP, a minimalist framework that is, to our knowledge, the first explicitly designed to suppress drift at its source within the PEFT-based FFT paradigm. OvA-LP combines linear probing on a frozen encoder with a one-vs-all head and a simple two-stage procedure, preserving pretrained feature geometry and decoupling logits to prevent the mechanisms that amplify drift. On CIFAR-100 with 100 clients, averaged over shard-1, shard-2, and Bernoulli--Dirichlet partitions, OvA-LP retains 95.9\% of its IID accuracy, whereas state-of-the-art FFT baselines retain only 10.1\% (PFPT) and 34.5\% (FFT-MoE) under the same conditions. OvA-LP further maintains resilience under both symmetric and asymmetric label noise. In addition, precomputing encoder features makes per-round cost nearly independent of encoder size. Together, these results demonstrate that OvA-LP provides a principled and efficient basis for robust FFT under heterogeneity.

\end{abstract}
\section{Introduction}

Foundation models (FMs) have reshaped machine learning by providing powerful pretrained representations that can be adapted to diverse downstream tasks. In federated learning (FL), this shift has given rise to federated fine-tuning (FFT), where clients adapt a shared encoder instead of training models from scratch \citep{zhuang2023foundation}. Parameter-efficient fine-tuning (PEFT) methods such as adapters, LoRA, and prompt tuning \citep{houlsby2019parameter, hu2022lora, lester2021power} further reduce computational and communication costs, making FFT a practical paradigm for large-scale decentralized adaptation. This paradigm is particularly critical in domains where data locality is a strict requirement, such as training on sensitive medical records across hospitals or financial data across institutions. However, despite its efficiency and practical importance, the robustness of FFT under heterogeneous client distributions remains a major challenge \citep{ren2025advances}. This gap motivates a rethinking of how to make FFT robust to extreme heterogeneity.

The core difficulty lies in local drift: client updates diverge due to distributional differences, and when aggregated, these drifts bias and destabilize the global model. Specifically, client-level divergences manifest as both systematic bias and amplified variance in the aggregated update, degrading final accuracy and hindering convergence. In practice, under extreme non-IID conditions, state-of-the-art FFT methods often converge slowly and retain well below half of their IID accuracy within 50 rounds. This persistent relative gap highlights the need for approaches that directly mitigate drift at the client level, rather than merely compensating for it after aggregation.

Broadly, prior efforts fall into two families. Aggregation strategies modify the global update rule, from classical methods such as FedProx and Scaffold \citep{li2020federated, karimireddy2020scaffold} to more recent LoRA-specific variants \citep{wang2024flora,guo2024selective,yanfederated}. Personalization frameworks attach client-specific modules to absorb drift locally, ranging from adapters and prompts to expert-based models \citep{zhang2024personalized,fan2024device,wangenhancing,yangfederated}. In practice these families are not disjoint—many approaches combine both, such as FFT-MoE\citep{hu2025fft} with expert routing on top of FedAvg, or PFPT \citep{weng2024probabilistic} with redesigned prompts and aggregation. Yet despite their variety, they share a common philosophy: drift is treated as unavoidable and corrected only post-hoc, once it has already manifested at the client or global level. This reactive stance leaves them fragile under extreme heterogeneity, as no method so far has succeeded in preventing drift from arising in the first place.

In this paper, we present OvA-LP, a minimalist framework that suppresses client drift at its root. 
OvA-LP integrates three lightweight components—linear probing (LP) on a frozen encoder \citep{alain2016understanding}, 
one-vs-all (OvA) binary heads, and a two-stage training schedule—each explored in isolation but never unified. 
By aligning them within a bias--variance decomposition of federated gradients, 
we systematically connect feature geometry, label decoupling, and variance control into a single source-level framework. 
This reframes drift mitigation from post-hoc correction to proactive prevention, 
suggesting shift in how robustness is pursued within PEFT-based FFT.

Our main findings are:
\begin{itemize}
    \item OvA-LP consistently prevents local drift from arising at the client level, offering a principled foundation for robust FFT under heterogeneity.
    \item OvA-LP retains 95.9\% of its IID accuracy on CIFAR-100 with 100 clients under shard-1, shard-2, and Bernoulli--Dirichlet ($p=0.1,\alpha=0.001$)\citep{xu2022fedcorr}, whereas FFT-MoE and PFPT retain only 34.5\% and 10.1\%, respectively.
    \item OvA-LP demonstrates innate robustness to label noise: it consistently reduces accuracy degradation under both symmetric and asymmetric corruption, maintaining resilience at higher noise levels and surpassing specialized noise-robust baselines.
    \item OvA-LP precomputes encoder features once, making per-round training nearly independent of encoder size and preserving modularity for integration with other FFT strategies.
\end{itemize}
\section{Related Work}

\paragraph{Aggregation strategies.}
A long line of work has sought to improve FL robustness by modifying the global update rule. Classical approaches such as FedProx and Scaffold reduce the variance of client updates and partially stabilize convergence. More recent extensions adapt these ideas to PEFT settings, for example FLoRA, FedSA-LoRA, and FRLoRA \citep{wang2024flora,guo2024selective,yanfederated}. While effective in mitigating some client drift, these methods still rely on aggregation at the server side, typically applied only after local divergence has already occurred.

\paragraph{Personalization frameworks.}
Another direction attaches client-specific modules to absorb drift locally. Examples include FedAdapter and FedPrompt \citep{cai2022fedadapter,zhao2023fedprompt}, as well as expert-based extensions such as FFT-MoE and PFPT. These approaches improve local adaptation, but global consistency remains limited because personalization cannot prevent drift from propagating into the shared model.

\paragraph{Classification heads for label imbalance.}
Another line of work modifies the classification head to mitigate skewed label distributions.  
FedRS \citep{li2021fedrs} restricts softmax updates for missing classes, mitigating bias under label imbalance.  
OvA-based approaches such as FedOVA, FedABC, and ATHENA-FL \citep{zhu2021fedova,wang2023fedabc,de2024athena} decompose multiclass tasks into binary classifiers to avoid softmax coupling and improve fairness.  
However, these methods are designed for scratch training and focus mainly on label imbalance, without addressing the broader challenge of feature drift.

\paragraph{Label noise robustness.}
A complementary line of research tackles noisy labels in FL. Methods such as FedCorr \citep{xu2022fedcorr} and FedLTF \citep{zhan2025fedltf} design correction mechanisms or robust objectives to improve performance under corruption. While effective, these approaches explicitly address noise rather than the underlying drift mechanisms, and remain orthogonal to our focus.

\paragraph{Our positioning.}
Unlike prior OvA-based methods restricted to scratch training and label imbalance, 
OvA-LP is designed to prevent drift from arising by freezing the encoder and introducing a two-stage OvA head.
Its minimalist design remains modular and in principle compatible with aggregation and personalization families, 
suggesting potential for deployment across diverse FFT pipelines.
\section{Methodology}
\begin{figure}[t]
    \centering
    \includegraphics[width=0.9\linewidth]{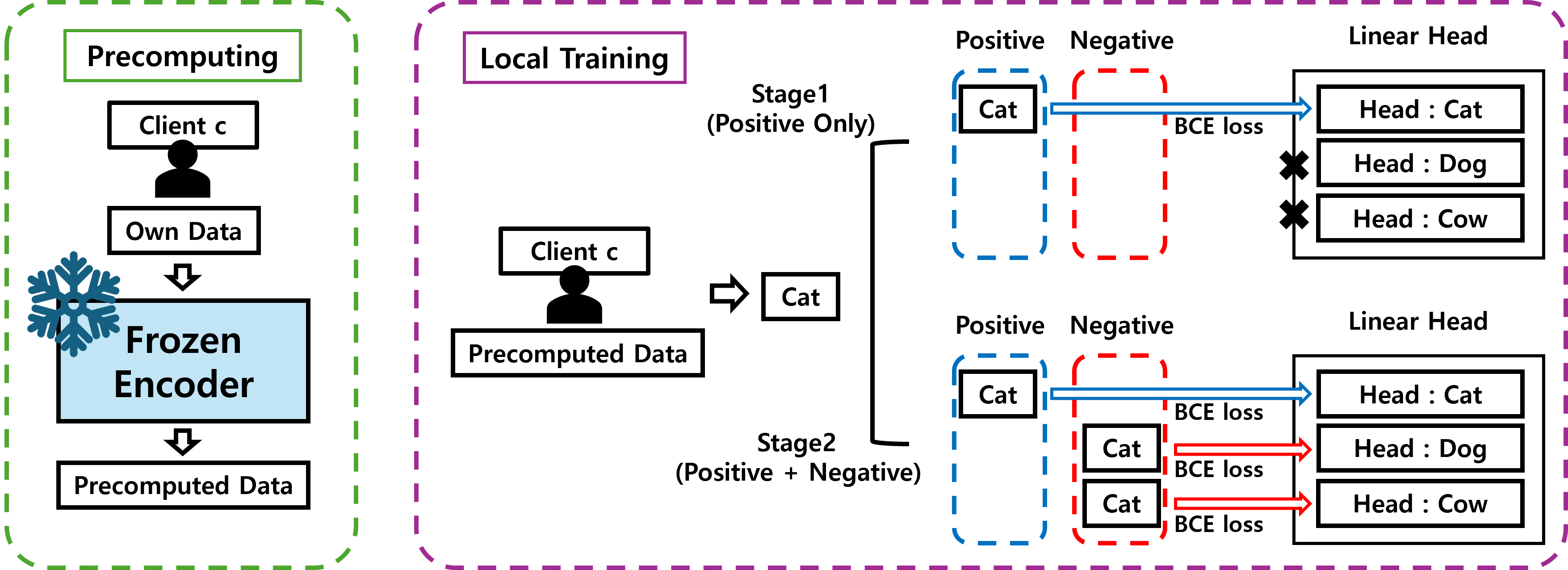}
    \caption{
        Overall structure of OvA-LP. Clients precompute encoder features once (left) 
        and perform two-stage local training with one-vs-all heads (right).
    }
    \label{fig:main}
\end{figure}

\paragraph{Overview.}
OvA-LP is motivated by a source-level philosophy: preventing drift at its origin rather than correcting it post hoc. 
Fig.~\ref{fig:main} summarizes the overall workflow. 
Clients first precompute encoder features with a frozen backbone, then train one-vs-all heads under a lightweight two-stage schedule. 
This design is guided by a bias--variance decomposition of federated gradients, 
which identifies local bias, global bias, and variance as the root causes of drift. 
OvA-LP targets each of these components with simple yet complementary mechanisms: 
feature geometry bounds the effect of feature skew, OvA heads eliminate label-skew bias and variance amplification, 
and the two-stage schedule stabilizes optimization under participation variance.  

The remainder of this section develops these ideas step by step. 
Sec.~\ref{sec:biasvariance} formalizes the bias--variance framework that motivates our design.  
Sec.~\ref{sec:feature} shows how pretrained geometry preserves alignment and separation, limiting bias from feature skew.  
Sec.~\ref{sec:ova} analyzes label skew, explaining how OvA decoupling removes the bias and variance amplification caused by softmax coupling.  
Finally, Sec.~\ref{sec:2stage} addresses the remaining variance, demonstrating how the two-stage schedule achieves fast and stable convergence.  
Together, these analyses show how OvA-LP systematically aligns with the bias--variance view to bring Non-IID training close to the IID reference.

\subsection{Bias--Variance Framework}
\label{sec:biasvariance}
We begin by formalizing drift through a bias--variance decomposition, 
which identifies local bias, global bias, and variance as the core sources of degradation.

Client drift under non-IID data can be understood through a bias–variance decomposition at both local and global levels. 
Let the stochastic gradient on client $i$ be $g_i = \nabla \ell(w; x,y)$. 
Denote by $\mathcal{D}_i$ the local data distribution of client $i$ and by $\mathcal{D}$ the global distribution. 
The local loss is $L_i(w) = \mathbb{E}_{(x,y) \sim \mathcal{D}_i}[\ell(w; x,y)]$ with expected gradient $\nabla L_i(w)$, 
and the global loss is $L(w) = \mathbb{E}_{(x,y) \sim \mathcal{D}}[\ell(w; x,y)]$ with gradient $\nabla L(w)$. 

\paragraph{Local bias.}
Each client’s optimum deviates from the global one by 
$b_i = \nabla L_i(w) - \nabla L(w)$, 
arising from distributional differences across clients, in particular feature skew and label skew.  

\paragraph{Global bias.}
Aggregating across clients yields 
$B = \mathbb{E}_i[\nabla L_i(w)] - \nabla L(w)$, 
which distorts the overall update direction and accumulates to reduce accuracy.

\paragraph{Local and global variance.}
Even within a single client, stochastic gradients fluctuate with variance $v_i = \mathrm{Var}[g_i]$. 
When aggregated with weights $p_i$ (e.g., proportional to dataset sizes $n_i$), the update is 
$\hat{g} = \sum_i p_i g_i$ with variance $V = \mathrm{Var}[\hat{g}]$, 
which is further amplified by quantity skew. 

\paragraph{Takeaway.}
Local bias and variance are the primary contributors to drift, while global bias and variance are their aggregated manifestations. 
Variance is further exacerbated under label skew due to softmax coupling, which introduces cross-class covariance. 
This explains why aggregation-level fixes cannot fundamentally solve non-IID degradation: they address only the aggregate symptoms rather than the underlying local causes.

\subsection{Linear Probing and Feature Geometry}

\label{sec:feature}
\begin{figure}[t]
    \centering
    \begin{minipage}{0.45\linewidth}
        \centering
        \small
        \resizebox{\linewidth}{!}{%
        \begin{tabular}{lcc}
        \toprule
        Metric & Pretrained (T) & Pretrained (F) \\
        \midrule
        Alignment $\downarrow$ & 1.366 $\pm$ 0.006 & 1.761 $\pm$ 0.008 \\ 
        Intra $\downarrow$     & 0.818 $\pm$ 0.002 & 0.917 $\pm$ 0.004 \\ 
        Inter $\uparrow$       & 0.840 $\pm$ 0.004 & 0.487 $\pm$ 0.008 \\ 
        Ratio $\downarrow$     & 0.974 $\pm$ 0.007 & 1.885 $\pm$ 0.038 \\ 
        \bottomrule
        \end{tabular}}
    \end{minipage}
    \hspace{4mm}
    \begin{minipage}{0.5\linewidth}
        \centering
        \includegraphics[width=\linewidth]{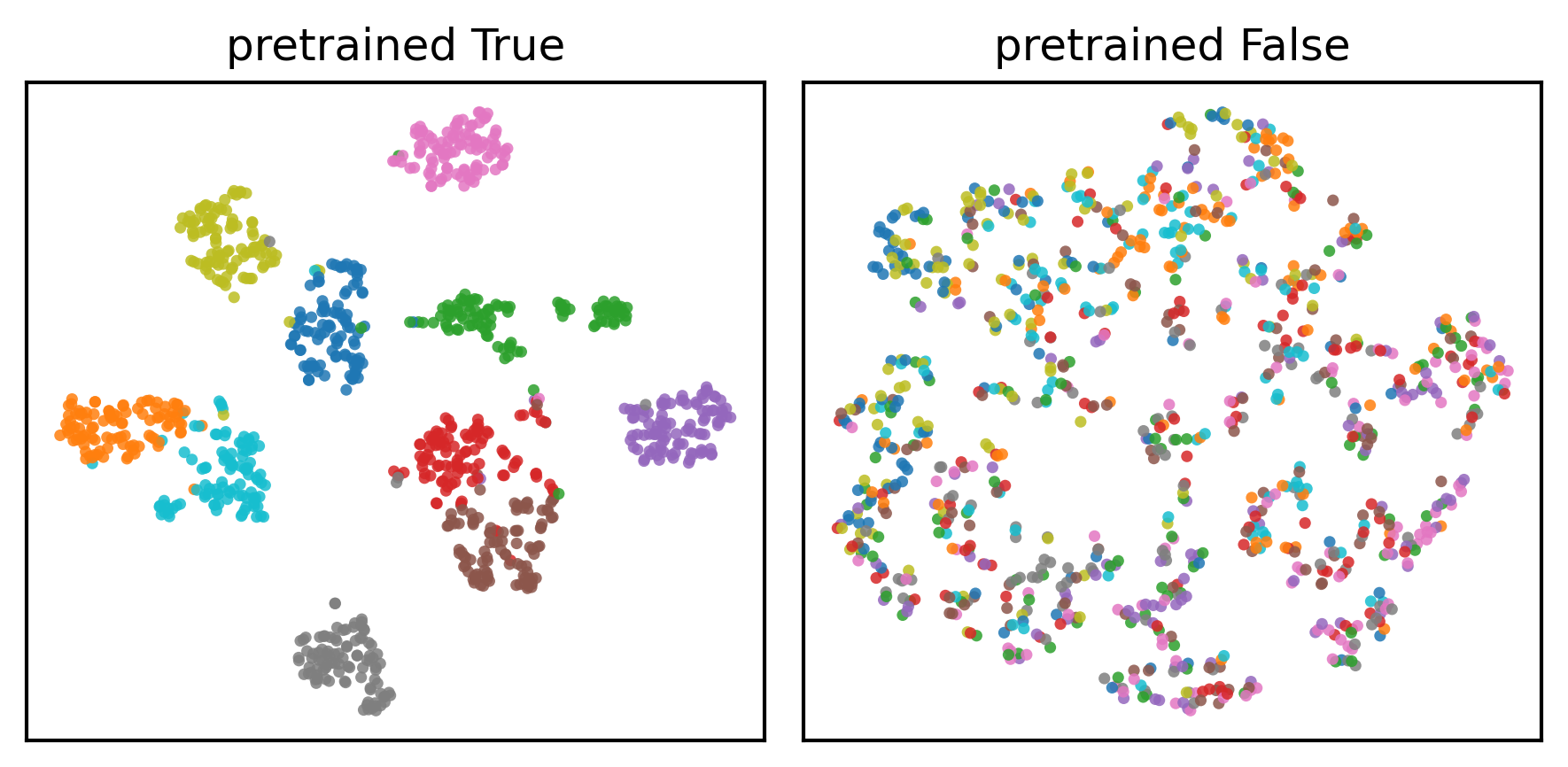}
    \end{minipage}
    \caption{Feature geometry of pretrained vs randomly initialized encoders (CIFAR-10, ViT-L/16).}
    \label{fig:feature-geometry-combined}
\end{figure}

Feature skew is bounded by pretrained geometry: alignment clusters same-class samples, 
and separation keeps classes apart.

We quantify feature geometry with four standard metrics. 
Following \citet{wang2020understanding}, alignment is defined as the expected squared distance between positive pairs:
\[
\text{Alignment} 
= \mathbb{E}_{(x,y)\sim p_{\text{pos}}} \, \| f(x) - f(y) \|_2^2.
\]

In addition, we report three well-known statistical measures of representation geometry:
\[
\text{Intra} 
= \mathbb{E}_{(x,y)} \, \| f(x) - \mu_y \|_2^2, 
\qquad
\text{Inter} 
= \mathbb{E}_{y \neq y'} \, \| \mu_y - \mu_{y'} \|_2^2,
\qquad
\text{Ratio} 
= \frac{\text{Intra}}{\text{Inter}}.
\]
Here $p_{\text{pos}}$ denotes the distribution over positive pairs, $\mu_y$ is the centroid of class $y$, 
and $f(\cdot)$ is the encoder representation. 
Alignment captures the closeness of positive pairs, Intra measures the compactness of each class cluster, 
Inter quantifies separation between class centroids, and Ratio summarizes the trade-off. 
Smaller Alignment, Intra, and Ratio and larger Inter indicate stronger feature geometry.

Fig.~\ref{fig:feature-geometry-combined} compares pretrained and randomly initialized encoders. 
Across all four metrics, pretrained features show smaller Alignment and Intra, larger Inter, and a lower Ratio, confirming that they form compact, well-separated clusters. 

From the bias--variance perspective, this structural geometry directly limits the bias induced by feature skew: 
alignment keeps same-class representations compact, while separation enforces clear boundaries across classes. 
As a result, client updates remain anchored to the global geometry, 
and the extent of local bias before aggregation is fundamentally bounded. 
In the ideal case of perfect alignment, feature-induced bias would vanish entirely.

\subsection{OvA Head and Decoupling}
\label{sec:ova}
The second major source of drift is label skew, which biases gradients and amplifies variance through softmax coupling.  
Let $h(x)=f_\theta(x)\in\mathbb{R}^d$ denote the encoder representation of input $x$, and let $w_c\in\mathbb{R}^d$ be the classifier weight vector for class $c$.  
We use $1[y=c]$ to denote the indicator for the ground-truth class.

\paragraph{Softmax coupling.}
For class $c$, the gradient of the cross-entropy loss with respect to $w_c$ is
\[
g_c(x,y) = (1[y=c] - p_c(x))h(x), \quad
p_c(x) = \frac{\exp(w_c^\top h(x))}{\sum_{j=1}^K \exp(w_j^\top h(x))}.
\]
Because all classes share a denominator, majority classes repeatedly dominate updates, 
while minority classes receive little signal. 
As analyzed in FedRS \citep{li2021fedrs}, this coupling introduces both bias and variance amplification under label skew. 
Replacing softmax with independent OvA heads removes this cross-class covariance, 
eliminating the mechanism behind label-skew drift. 
As analyzed in FedRS \citep{li2021fedrs}, majority classes dominate through repeated ``pulls,'' while minority classes often receive only ``pushes.''  
This imbalance introduces bias, since updates are driven by probability-weighted terms $p_c(x)$ rather than purely class-specific targets.  
It also amplifies variance, because the shared denominator induces non-zero cross-class covariances $\mathrm{Cov}(g_c,g_j)\neq 0$.  
Together, these mechanisms destabilize training under heterogeneous distributions.

\paragraph{OvA decoupling.}
An OvA head replaces softmax with independent binary classifiers.  
The gradient of the logistic loss with respect to $w_c$ is
\[
g^{\text{OvA}}_c(x,y) = (1[y=c] - q_c(x))h(x), \quad 
q_c(x) = \sigma(w_c^\top h(x)) = \tfrac{1}{1+\exp(-w_c^\top h(x))}.
\]
$q_c(x)$ is the Bernoulli likelihood under a logistic regression head, and each head optimizes its binary logistic loss independently.  
As a result, the pull/push imbalance described in FedRS disappears: majority and minority classes are updated without mutual interference.  
This decoupling eliminates the mechanism of label-skew-induced bias and variance amplification, directly addressing the sources of drift at their origin.

\subsection{Variance and Two-Stage Training}
\label{sec:2stage}
After bias terms are suppressed, variance remains the main source of drift.
Variance cannot be eliminated entirely, but its destabilizing effect can be controlled through a two-stage curriculum aligned with the OvA structure.

\paragraph{Stage 1 (positive-only).}
When pretrained representations preserve alignment and separation, 
the global optimum of each OvA head lies near the class centroid at the point of maximum margin. 
Training only on positives thus pulls classifier weights toward these centroids, 
leading to rapid convergence without cross-class conflicts and helping to 
overcome the destabilizing effect of variance in the early rounds.

\paragraph{Stage 2 (positive+negative).}
After centroids are established, a large set of negatives is introduced to expand inter-class margins. 
At the same time, a small fraction of positives is retained as anchors, preventing the decision boundary from drifting away under the stronger influence of negatives. 
This combination enables efficient margin learning while preserving the stability achieved in Stage~1.

\paragraph{Takeaway.}
Together, the two stages implement an easy-first, hard-later curriculum. 
Stage~1 quickly aligns classifiers with class centroids under minimal variance, while Stage~2 leverages negatives for margin expansion without destabilizing the positive clusters. 
This design directly overcomes variance effects at their source, complementing OvA-LP’s treatment of feature and label skew.

\section{Experiments}

\subsection{Experimental Setup}

\paragraph{Shared setting.}
\label{sec:setup}
Our primary experiments use CIFAR-100 with 100 clients for 50 rounds, a scale comparable to or larger than those adopted in recent FL benchmarks (see Appendix~\ref{app:baseline_dataset} for survey). 
We fix five random seeds (0, 42, 777, 1337, 15254) across all runs for comparability. 
For the IID setting, data are split uniformly at random across clients. 
For the Non-IID setting, we use three representative configurations widely adopted in the literature: 
Shard-1 (one class per client), 
Shard-2 (two classes per client), 
and Dirichlet ($p=0.1,\alpha=0.001$) following the FedCorr construction~\citep{xu2022fedcorr}. 
These serve as the standard Non-IID benchmarks throughout. 
Further analyses in Sec.~4.4 expand beyond this shared setting, including alternative partitions, encoder scaling, datasets like TinyImageNet, and robustness to label noise.

\paragraph{Our model.}
OvA-LP uses a frozen ViT-L/16 encoder and is trained with 100\% client participation, three local epochs per round, batch size 50, learning rate 0.01, and AdamW optimizer with weight decay $1\times10^{-4}$. 
Client updates are aggregated by FedAvg \citep{mcmahan2017communication}, with the first round conducted using Stage~1 (positive-only) training and all subsequent rounds using Stage~2 (positive+negative) training, as described in Sec.~3.4. 

\paragraph{Baselines.}
Baseline methods are reproduced using their original model architectures and training protocols as specified in their papers. 
We preserve the encoders used in the original implementations (e.g., ViT-B/32, ViT-B/16), ensuring faithful reproduction; detailed configurations are reported in Appendix~\ref{app:baseline_parameters}. 

\paragraph{Evaluation philosophy.}
We quantify non-IID robustness by comparing accuracy trajectories to the IID reference. 
For round $t$, we compute the relative ratio $R(t) = \text{Acc}_{\text{NonIID}}(t) / \text{Acc}_{\text{IID}}(t) \times 100$. 
We present results in two unified views: round-wise $R(t)$ curves showing how quickly and stably each method tracks the IID trajectory, and final $R(50)$ barplots summarizing the endpoint gap for each partition. 
This framing provides a consistent lens that captures convergence speed, stability, and final accuracy.

\subsection{Ablation Study}
\label{sec:ablation}

We examine the contribution of each design component of OvA-LP by comparing three head configurations: 
(i) LP with softmax, 
(ii) OvA-LP without the two-stage design, 
and (iii) the full OvA-LP. 

\begin{figure}[h]
    \centering
    \includegraphics[width=0.9\textwidth]{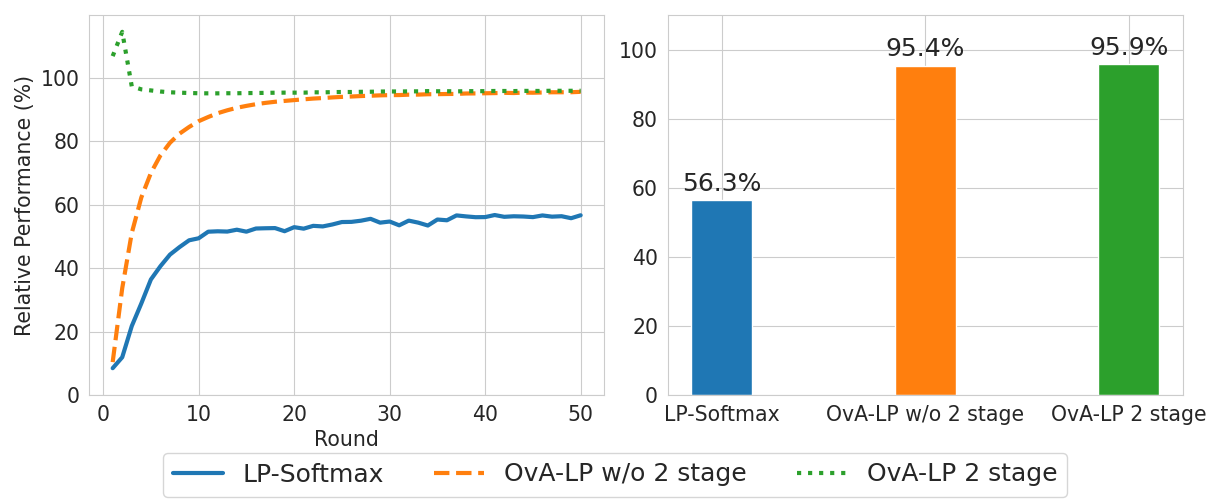}
    \caption{Ablation of OvA-LP components. 
    Stepwise gains (56.3 $\rightarrow$ 95.4 $\rightarrow$ 95.9) illustrate the effects of OvA decoupling and two-stage training.}
    \label{fig:ablation}
\end{figure}

As shown in Fig.~\ref{fig:ablation}, the progression is stepwise. 
LP-softmax reaches only 56.3\% under Non-IID, reflecting limited benefit from encoder freezing alone.  
Replacing the softmax with independent OvA classifiers stabilizes training and raises performance to 95.4\%.  
Adding the two-stage design enables faster convergence to 95.9\%, 
closely tracking the IID curve within only a few rounds.

A brief overshoot above the IID curve occurs in the first few rounds. 
This behavior arises from FedAvg’s weighted averaging under imbalanced partitions and quickly settles.  

In summary, OvA decoupling mitigates label-skew effects, and the two-stage procedure helps overcome variance, 
leading to faster and more stable convergence. 
These observations align with the bias--variance decomposition described in Sec.~\ref{sec:2stage}.  
Full accuracy curves and efficiency breakdowns are reported in Appendix~\ref{app:ablation}.

\subsection{Comparison with Baselines}

We next compare OvA-LP against two recent state-of-the-art baselines, FFT-MoE and PFPT. 

\begin{figure}[t]
    \centering
    \includegraphics[width=0.9\textwidth]{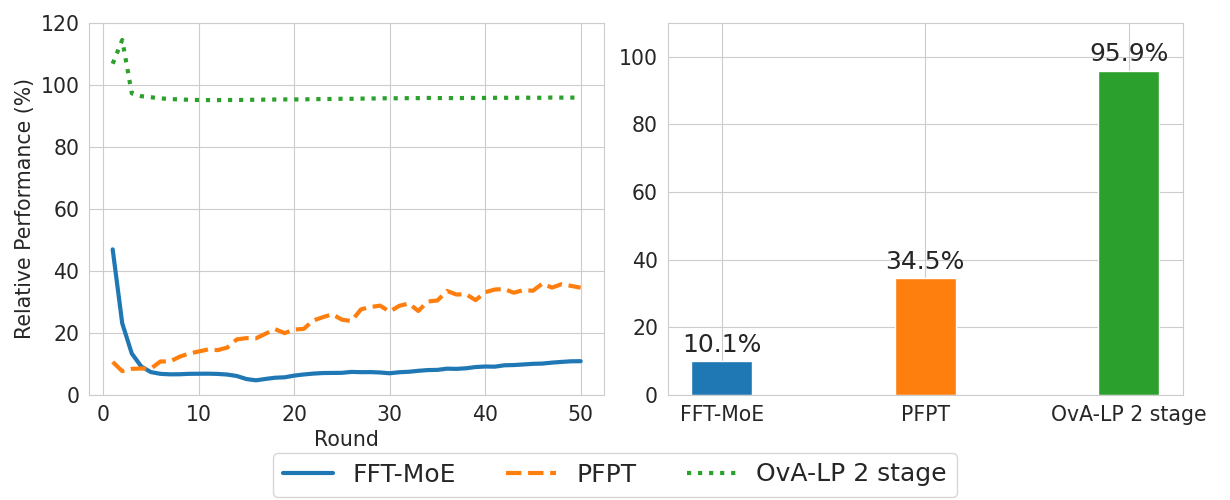}
    \caption{Comparison with state-of-the-art baselines. 
    FFT-MoE plateaus near 10.1\%, while PFPT rises slowly but saturates at 34.5\%. 
    OvA-LP remains stable and converges to 95.9\%.}
    \label{fig:baseline}
\end{figure}

Fig.~\ref{fig:baseline} shows the outcome under the extreme Non-IID setting.  
Both baselines perform poorly compared to the IID reference: PFPT increases gradually but saturates at 34.5\%, 
while FFT-MoE plateaus early and remains near 10.1\%.  
These results are consistent with their post-hoc philosophy, which seeks to mitigate drift only after aggregation.  

OvA-LP, in contrast, maintains stability and converges to 95.9\%, closely following the IID trajectory.  
As noted in Sec.~\ref{sec:ablation}, even LP-softmax, which partially reduces feature bias through encoder freezing, 
already surpasses post-hoc baselines in this setting.  
OvA-LP further improves upon this by also addressing label skew and variance effects, 
leading to accuracy near the IID level.  
Additional comparisons with FFT-MoE and PFPT are provided in Appendix~\ref{app:baseline_cmp}.  

\begin{table}[t]
\centering
\begin{tabular}{lccc c}
\toprule
Method & Label Bias & Feature Bias & Var. & $R(50) (\%)$ \\
\midrule
FFT-MoE / PFPT & $\times$ & $\times$ & $\times$ & 10.1 / 34.5 \\
LP-softmax & $\times$ & $\triangle$ & $\times$ & 56.3 \\
OvA-LP (w/o 2-stage) & \checkmark & $\triangle$ & $\times$ & 95.4 \\
OvA-LP (2-stage) & \checkmark & $\triangle$ & \checkmark & 95.9 \\
\bottomrule
\end{tabular}
\caption{Bias–variance view of methods. 
“$\checkmark$” = removed/handled, “$\triangle$” = partially removed, “$\times$” = not addressed. 
As non-IID severity increases, leaving label bias intact results in low robustness, 
while OvA-LP progressively removes label bias and handles variance to reach near-IID performance.}
\label{tab:biasvariance}
\end{table}

The ranking across methods follows a stepwise pattern.  
Post-hoc baselines leave both label and feature bias intact, resulting in poor robustness under strong heterogeneity.  
LP-softmax reduces feature bias but retains label bias, leading to moderate accuracy.  
OvA-LP without the two-stage procedure removes label bias and improves stability, 
and the full OvA-LP additionally addresses variance, reaching near-IID robustness.  
This stepwise progression aligns with the bias--variance decomposition and illustrates the benefit of addressing drift at its origin, as summarized in Table~\ref{tab:biasvariance}.

\subsection{Additional Analyses}

\subsubsection{Partition-wise Robustness.}

\begin{figure}[t]
    \centering 
    \includegraphics[width=0.9\textwidth]{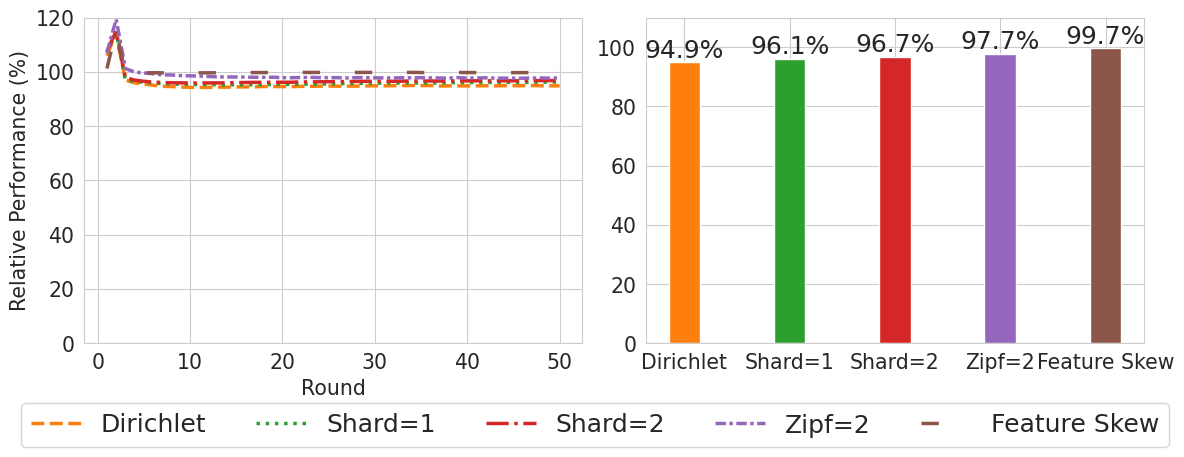}
    \caption{Partition-wise robustness of OvA-LP. 
    Across five representative heterogeneity patterns, 
    $R(t)$ curves (left) closely track the IID reference and final $R(50)$ values (right) remain above 94.9\%. 
    This confirms consistent robustness across diverse forms of skew, including label, feature, and quantity heterogeneity.}
    \label{fig:partition}
\end{figure}

We evaluate OvA-LP under five representative heterogeneity patterns. 
Three of them—Shard-1, Shard-2, and Dirichlet ($\alpha=0.001, p=0.1$)—are the standard benchmarks already introduced in Sec.~\ref{sec:setup}. 
We further include two additional settings. 
First, we adopt a Zipf distribution with exponent $s=2.0$, a standard setup in FL for inducing quantity skew across clients~\citep{piantadosi2014zipf}.
Second, feature-based clustering, where each class is partitioned into $K$ clusters (with $K$ equal to the number of clients) using $k$-means, and each cluster is then assigned to a client.

Fig.~\ref{fig:partition} shows that across all five settings, the $R(t)$ curves remain aligned with the IID trajectory, 
and final $R(50)$ values range from 94.9\% to 99.7\%. 
This demonstrates that OvA-LP maintains robustness under diverse forms of skew, including label, feature, and quantity heterogeneity.

\subsubsection{Encoder and Task Variations.}

\begin{figure}[t]
    \centering
    \includegraphics[width=0.9\textwidth]{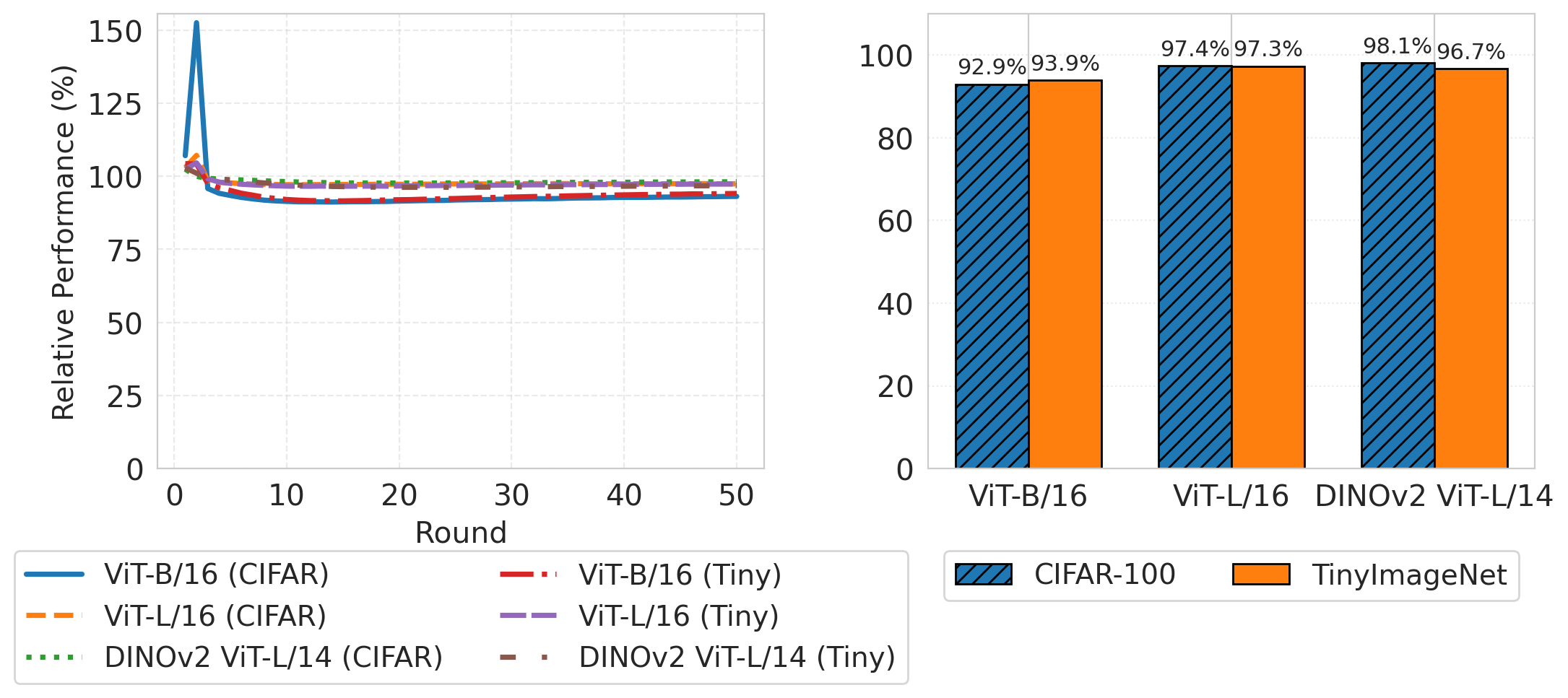}
    \caption{Encoder and task variations. OvA-LP is evaluated with different encoders (ViT-B/16, ViT-L/16, DINOv2-L/14) on CIFAR-100 and extended to TinyImageNet under Dirichlet partitioning.}
    \label{fig:encoder_dataset}
\end{figure}

We next test whether robustness depends on encoder scale or task domain. 
Fig.~\ref{fig:encoder_dataset} compares ViT-B/16, ViT-L/16, and DINOv2-L/14 on CIFAR-100, and extends to TinyImageNet under Dirichlet partitioning. 
Absolute accuracy decreases for the smallest encoder (ViT-B/16), while ViT-L/16 and DINOv2-L/14 remain comparable, with minor fluctuations depending on the dataset. 
Crucially, in all cases $R(t)$ curves consistently track the IID trajectory, confirming that OvA-LP’s robustness is agnostic to encoder scale, architecture, and task domain. 

We note that the brief overshoot observed in the first round is a benign effect of size-weighted FedAvg, 
which appears more prominently with smaller encoders. 
It stabilizes quickly and does not affect final convergence.

\subsubsection{Label Noise Robustness.}

\begin{table*}[t]
    \centering
    \resizebox{\textwidth}{!}{%
    
    \begin{tabular}{l c cccc c cc}
        \toprule
        Noise Type & \multicolumn{5}{c}{Symmetric} & \multicolumn{3}{c}{Asymmetric} \\
        \midrule
        Method & Baseline Acc (\%) & \multicolumn{4}{c}{Decline Rate(\%) $\downarrow$} & Baseline Acc (\%) & \multicolumn{2}{c}{Decline Rate(\%) $\downarrow$} \\
        \midrule
        Noise Ratio & 0.30 & 0.40 & 0.50 & 0.60 & 0.70 & 0.20 & 0.30 & 0.40 \\
        \midrule
        FedAvg & 16.75\% & 15.70\% & 23.70\% & 37.49\% & 51.46\% & 18.85\% & 13.37\% & 30.93\% \\        
        Symmetric CE & 16.99\% & 17.77\% & 25.66\% & 40.79\% & 49.79\% & 26.14\% & 17.71\% & 36.34\% \\
        Co-teaching & 34.21\% & 8.68\% & 36.07\% & 51.04\% & 66.99\% & 34.19\% & 20.10\% & 33.23\% \\
        FedCorr & 32.15\% & 13.50\% & 26.59\% & 41.65\% & 62.64\% & 41.12\% & 13.47\% & 30.81\% \\
        FedNoRo & 38.58\% & 9.46\% & 19.57\% & 35.38\% & 43.36\% & 45.42\% & 9.82\% & 26.97\% \\
        FedLTF (Stage 2) & 55.23\% & 3.73\% & 8.73\% & 12.18\% & 21.24\% & 52.63\% & 10.94\% & 26.51\% \\
        FedLTF (Stage 3) & 58.43\% & 3.70\% & 9.24\% & 14.65\% & 20.91\% & 57.78\% & 8.71\% & 24.80\% \\
        \midrule
        OvA-LP (Ours)    & 88.78\% & \textbf{0.76\%} & \textbf{2.35\%} & \textbf{4.52\%} & \textbf{10.35\%} & 89.28\% & \textbf{0.63\%} & \textbf{1.53\%} \\
        \bottomrule
    \end{tabular}
    }
    \caption{Robustness on CIFAR-100 with label noise, measured as accuracy decline rates (\%) from baseline accuracy. 
    Baseline results are taken from FedLTF~\citep{zhan2025fedltf} (Table 2), which included standard and robust training schemes as well as its own variants. 
    OvA-LP (2-stage) achieves the smallest decline, outperforming the prior state-of-the-art FedLTF.}
    \label{tab:noise_decay}
\end{table*}

Following FedLTF~\citep{zhan2025fedltf}, we adopt the same label noise benchmarks and directly compare against the baselines it reports. 
FedLTF represented the prior state-of-the-art under label corruption. 
As Table~\ref{tab:noise_decay} shows, OvA-LP achieves markedly smaller accuracy declines, surpassing FedLTF and all other reported methods.

\paragraph{Summary.}
Taken together, these analyses show that OvA-LP retains stability under diverse forms of heterogeneity, across encoder scales and task domains, and even in the presence of label corruption, demonstrating robustness across a broad range of conditions.

\section{Limitations}
\label{sec:limitations}
We note two main limitations. 
First, OvA-LP relies heavily on the pretrained encoder: alignment and separation in the encoder’s feature geometry are what reduce feature-skew bias and enable linear probing. 
If the encoder is weak, OvA-LP cannot compensate on its own. 
This is not unique to our method but reflects the broader trend in federated fine-tuning, where progress is fundamentally tied to advances in foundation models. 

Second, all experiments assume full client participation. 
This choice highlights fast convergence under minimal variance but abstracts away from partial participation, which is common in practice. 
Indeed, as shown in Appendix~\ref{app:participation}, reduced participation slows convergence, 
and our two-stage strategy alone cannot fully overcome this variance. 
However, Appendix~\ref{app:baseline_cmp} demonstrates that OvA-LP remains highly efficient under full participation: 
it reaches Acc@95 within only 1–3 rounds, with both computation and communication costs substantially lower than prior methods, 
even when all 100 clients are active. 
Thus, while partial participation exposes a limitation, 
the lightweight design of OvA-LP makes full participation not only operationally feasible but also a practical advantage in real deployments.

In addition, our study is limited to vision benchmarks and does not yet combine with aggregation or personalization frameworks. 
We regard these as natural directions for future work.

\section{Conclusion}

We introduced OvA-LP, a minimalist framework for federated fine-tuning that addresses client drift at its origin. 
By combining linear probing with a one-vs-all head and a simple two-stage training strategy, OvA-LP shows that non-IID robustness can be achieved without architectural complexity. 
Despite its simplicity, it reaches near-IID accuracy across a wide range of non-IID settings, exhibits robustness to label noise consistent with its bias–variance suppression design, and maintains efficiency that scales favorably with encoder size. 

Our perspective does not dismiss existing aggregation or personalization strategies; rather, it offers a complementary direction. 
Where prior approaches mitigate drift after it emerges, OvA-LP prevents its amplification at the source, making it especially effective under extreme heterogeneity. 
At the same time, post-hoc methods remain valuable for personalization and fine-grained corrections, suggesting a natural synergy with our framework. 

OvA-LP stands as a strong new baseline and an initial step toward making source-level robustness a standard paradigm in federated fine-tuning.

\section*{Acknowledgements}
This research was supported by the Institute of Information  Communications Technology Planning \& Evaluation (IITP)
grant funded by the Korea government (MSIT)
(No.RS-2024-00398360, Development of a userfriendly and efficiency-optimized real-time homomorphic statistical analysis processing platform \& No.RS-2024-00399491, Development of Privacy-Preserving Multiparty Computation Techniques for Secure Multiparty Data Integration)

\bibliographystyle{unsrtnat}
\bibliography{references}

\appendix
\section{Additional Experimental Results}

\subsection{Ablation Results}
\label{app:ablation}

\begin{figure}[h!]
    \centering 
    \includegraphics[width=0.8\textwidth]{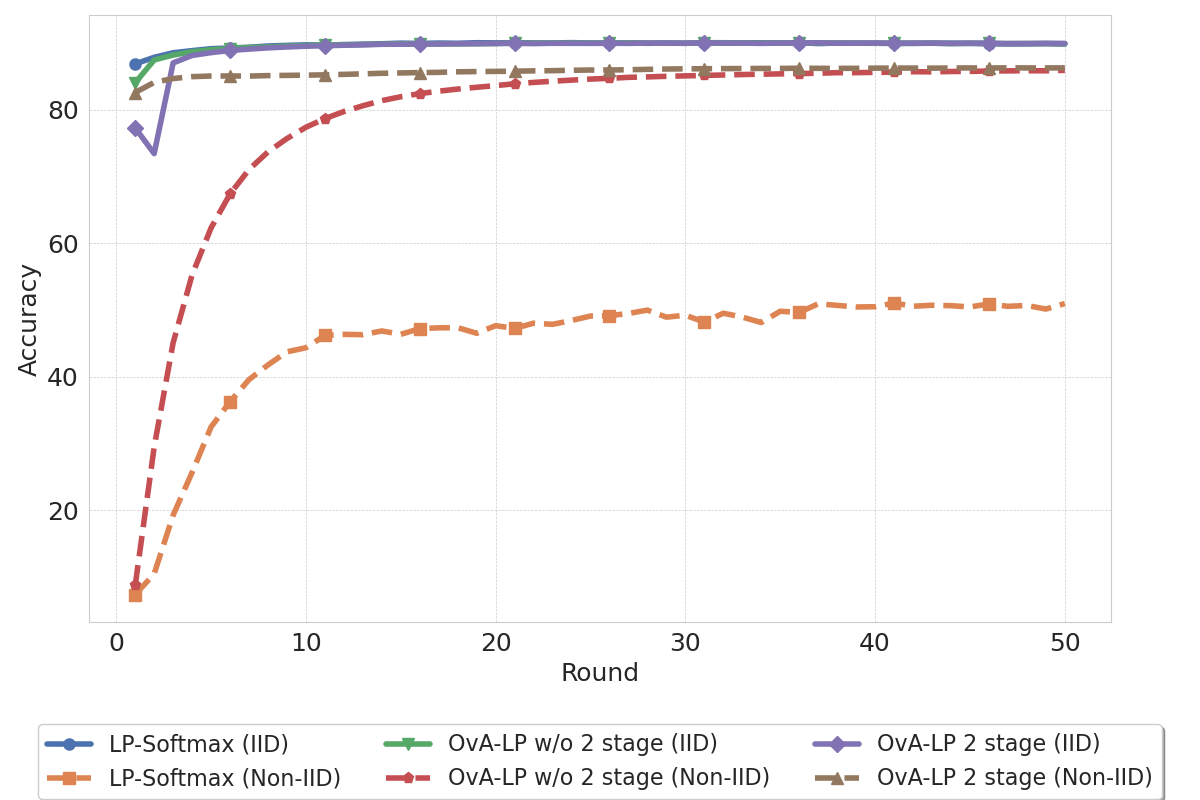} 
    \caption{Ablation curves under IID and averaged Non-IID settings. 
    Accuracy trajectories over 50 rounds.}
    \label{fig:app_ablation}
\end{figure}

\begin{table*}[h!]
\centering
\label{tab:main_performance}
\resizebox{0.9\textwidth}{!}{%
\begin{tabular}{@{}lcccc@{}}
\toprule
\textbf{Methodology} & \textbf{Accuracy (\%)} & \textbf{Acc@95 (Rounds)} & \textbf{Total Time (s)} & \textbf{Total Comm. (MB)} \\
\midrule
LP-Softmax (IID) & $90.13 \pm 0.04$ & $1 \pm 0$ & $0.03 \pm 0.00$ & $0.39 \pm 0.00$ \\
LP-Softmax (Non-IID) & $53.70 \pm 6.79$ & $27 \pm 9$ & $0.73 \pm 0.23$ & $10.71 \pm 3.37$ \\
\midrule
OvA-LP w/o 2 stage (IID) & $90.03 \pm 0.08$ & $2 \pm 0$ & $0.05 \pm 0.00$ & $0.78 \pm 0.00$ \\
OvA-LP w/o 2 stage (Non-IID) & $85.89 \pm 1.11$ & $15 \pm 3$ & $0.37 \pm 0.07$ & $5.72 \pm 1.02$ \\
\midrule
OvA-LP 2 stage (IID) & $90.04 \pm 0.12$ & $3 \pm 0$ & $0.05 \pm 0.00$ & $1.18 \pm 0.00$ \\
OvA-LP 2 stage (Non-IID) & $86.34 \pm 0.73$ & $1 \pm 0$ & $0.02 \pm 0.00$ & $0.42 \pm 0.10$ \\
\bottomrule
\end{tabular}%
}
\caption{Final performance metrics of ablation study, including accuracy, convergence rounds (Acc@95), total time, and total communication until convergence.}
\label{tab:app_ablation1}
\end{table*}


\begin{table*}[h!]
\centering
\label{tab:resource_metrics}
\resizebox{0.9\textwidth}{!}{%
\begin{tabular}{@{}lcccc@{}}
\toprule
\textbf{Methodology} & \textbf{Time (Client, ms)} & \textbf{Time (Server, ms)} & \textbf{Comm. (Client)} & \textbf{Comm. (Server)} \\
\midrule
LP-Softmax (IID) & $23.18 \pm 0.17$ & $2.64 \pm 0.04$ & 401.20 KB & 39.18 MB \\
LP-Softmax (Non-IID) & $23.78 \pm 0.60$ & $2.97 \pm 0.11$ & 401.20 KB & 39.18 MB \\
\midrule
OvA-LP w/o 2 stage (IID) & $21.67 \pm 0.21$ & $2.69 \pm 0.06$ & 401.20 KB & 39.18 MB \\
OvA-LP w/o 2 stage (Non-IID) & $22.27 \pm 0.48$ & $2.99 \pm 0.03$ & 401.20 KB & 39.18 MB \\
\midrule
OvA-LP 2 stage (IID) & $14.15 \pm 0.22$ & $1.90 \pm 0.04$ & 401.20 KB & 39.18 MB \\
OvA-LP 2 stage (Non-IID) & $14.86 \pm 0.51$ & $1.91 \pm 0.06$ & 401.20 KB & 39.18 MB \\
\bottomrule
\end{tabular}%
}
\caption{Per-round computation and communication costs of ablation study.}
\label{tab:app_ablation2}
\end{table*}

\noindent
Figure~\ref{fig:app_ablation} presents the detailed accuracy results from the ablation study.
In contrast to the IID vs. Non-IID comparison graphs in the main text, it reports the absolute accuracies for both settings separately.

Tables~\ref{tab:app_ablation1} and \ref{tab:app_ablation2} summarize convergence and per-round costs.  
Under IID, all methods converge quickly with small differences; OvA-LP (2-stage) shows a slight increase in communication due to additional heads.  

The differences appear under Non-IID: LP-Softmax requires 27 rounds and about 0.73s to reach Acc@95, 
whereas OvA-LP (2-stage) reaches the same point in a single round, reducing total time by $\sim$36$\times$ and communication by $\sim$25$\times$.  
Per-round metrics show that OvA-LP (2-stage) is also more efficient, requiring 14.9ms on clients versus 23.8ms for LP-Softmax, and 1.9ms on the server versus 3.0ms.  

In summary, the ablation indicates that while all methods behave similarly under IID, 
under Non-IID the full OvA-LP achieves near-IID efficiency with substantially fewer rounds and lower system cost.

\subsection{Baseline Comparisons}
\label{app:baseline_cmp}

\begin{figure}[h!]
    \centering
    \includegraphics[width=0.8\textwidth]{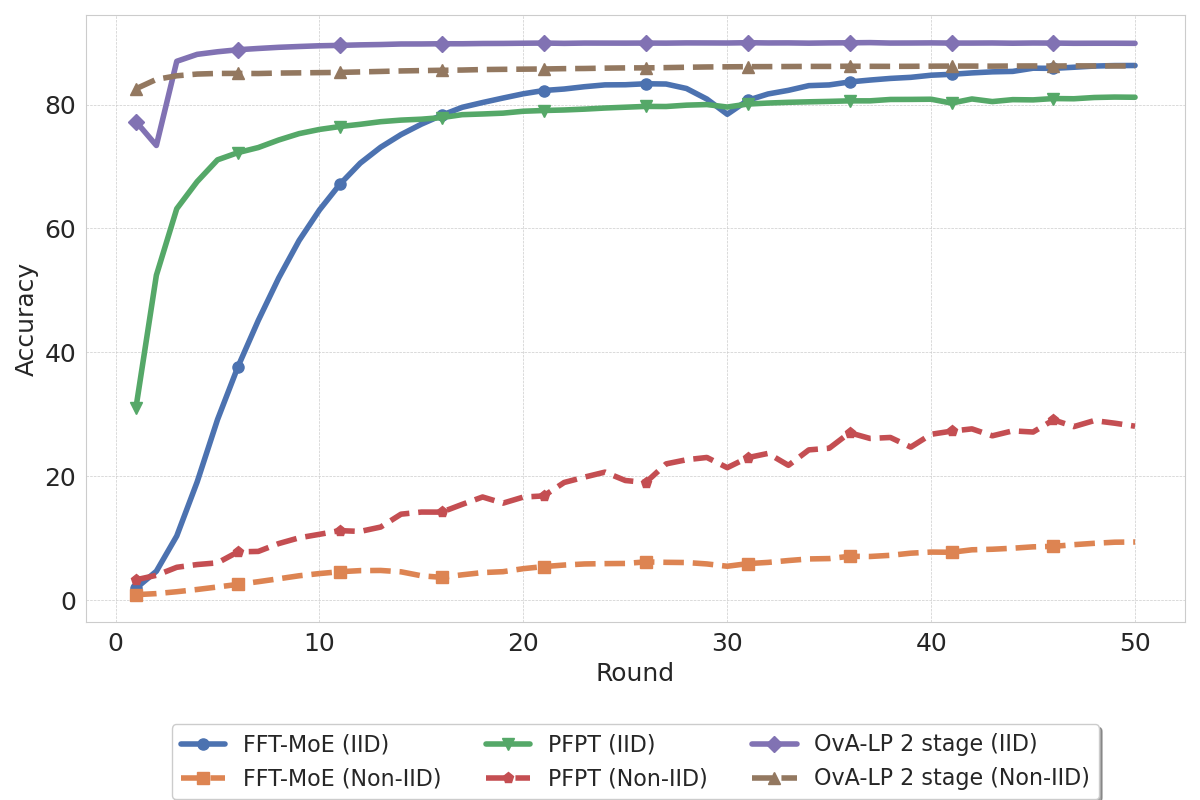}
    \caption{Baseline curves under IID and averaged Non-IID settings. 
    Accuracy trajectories over 50 rounds.}
    \label{fig:app_baseline}
\end{figure}

\noindent
Figure~\ref{fig:app_baseline} presents the actual accuracy results from the baseline study.
Unlike Figure~\ref{fig:app_ablation}, which reports the outcomes of the ablation study, this figure shows the absolute accuracies obtained from baseline experiments under both IID and Non-IID settings.

\begin{table*}[h!]
\centering
\resizebox{0.9\textwidth}{!}{
\begin{tabular}{@{}lcccc@{}}
\toprule
\textbf{Methodology} & \textbf{Accuracy (\%)} & \textbf{Acc@95 (Rounds)} & \textbf{Total Time (s)} & \textbf{Total Comm. (GB)} \\
\midrule
FFT-MoE (IID) & $96.39 \pm 0.41$ & $21 \pm 0$ & \textbf{$34.58 \pm 0.70$} & \textbf{$3.67 \pm 0.08$} \\
FFT-MoE (Non-IID) & $9.83 \pm 7.96$ & $37 \pm 13$ & \textbf{$61.00 \pm 21.59$} & \textbf{$6.48 \pm 2.32$} \\
\midrule
PFPT (IID) & $80.27 \pm 0.41$ & $13 \pm 1$ & \textbf{$432.75 \pm 32.45$} & \textbf{$0.22 \pm 0.02$} \\
PFPT (Non-IID) & $33.17 \pm 15.30$ & $44 \pm 6$ & \textbf{$1437.00 \pm 182.03$} & \textbf{$0.74 \pm 0.09$} \\
\midrule
OvA-LP (IID) & $90.04 \pm 0.12$ & $3 \pm 0$ & \textbf{$0.05 \pm 0.00$} & \textbf{$0.11 \pm 0.00$} \\
OvA-LP (Non-IID) & $86.34 \pm 0.73$ & $1 \pm 0$ & \textbf{$0.02 \pm 0.00$} & \textbf{$0.04 \pm 0.01$} \\
\bottomrule
\end{tabular}
}
\caption{Final performance of baselines: accuracy, convergence rounds (Acc@95), and total costs until convergence.}
\label{tab:app_baseline1}
\end{table*}

\begin{table*}[h!]
\centering
\resizebox{0.9\textwidth}{!}{
\begin{tabular}{@{}lcccc@{}}
\toprule
\textbf{Methodology} & \textbf{Time (Client, ms)} & \textbf{Time (Server, ms)} & \textbf{Comm. (Client)} & \textbf{Comm. (Server)} \\
\midrule
FFT-MoE (IID) & $1560.27 \pm 10.38$ & $94.76 \pm 0.42$ & 1.7703 MB & 177.03 MB \\
FFT-MoE (Non-IID) & $1555.74 \pm 18.46$ & $95.49 \pm 1.28$ & 1.7703 MB & 177.03 MB \\
\midrule
PFPT (IID) & $1860.98 \pm 10.01$ & $30941.43 \pm 387.65$ & 1.729 MB & 17.29 MB \\
PFPT (Non-IID) & $1855.49 \pm 39.89$ & $31002.22 \pm 239.84$ & 1.729 MB & 17.29 MB \\
\midrule
OvA-LP (IID) & $14.16 \pm 0.22$ & $1.90 \pm 0.04$ & 0.3918 MB & 39.18 MB \\
OvA-LP (Non-IID) & $14.86 \pm 0.51$ & $1.91 \pm 0.06$ & 0.3918 MB & 39.18 MB \\
\bottomrule
\end{tabular}
}
\caption{Per-round computation and communication costs of baselines.}
\label{tab:app_baseline2}
\end{table*}

\noindent
Tables~\ref{tab:app_baseline1} and \ref{tab:app_baseline2} compare FFT-MoE, PFPT, and OvA-LP under both IID and Non-IID.  
FFT-MoE achieves strong accuracy under IID (96\%) but collapses under Non-IID, converging below 10\% even after 37 rounds.  
PFPT is more stable across settings but converges slowly: its time-to-95\% accuracy exceeds OvA-LP by over three orders of magnitude, despite using less communication.  
In contrast, OvA-LP converges within 3 rounds (IID) and 1 round (Non-IID), while its final accuracy remains above 86--90\%.  
This corresponds to $10^2$–$10^4$ reductions in time and communication compared to prior baselines.  
Per-round metrics further confirm the gap: OvA-LP requires only $\sim$14 ms on clients and 2 ms on the server, 
versus seconds or tens of seconds for PFPT and FFT-MoE.  

Overall, OvA-LP attains comparable or better final accuracy while reaching convergence substantially faster and at far lower system cost.

\subsection{Participation Rate Sweep}
\label{app:participation}

\begin{figure}[h!]
    \centering
    \includegraphics[width=0.95\textwidth]{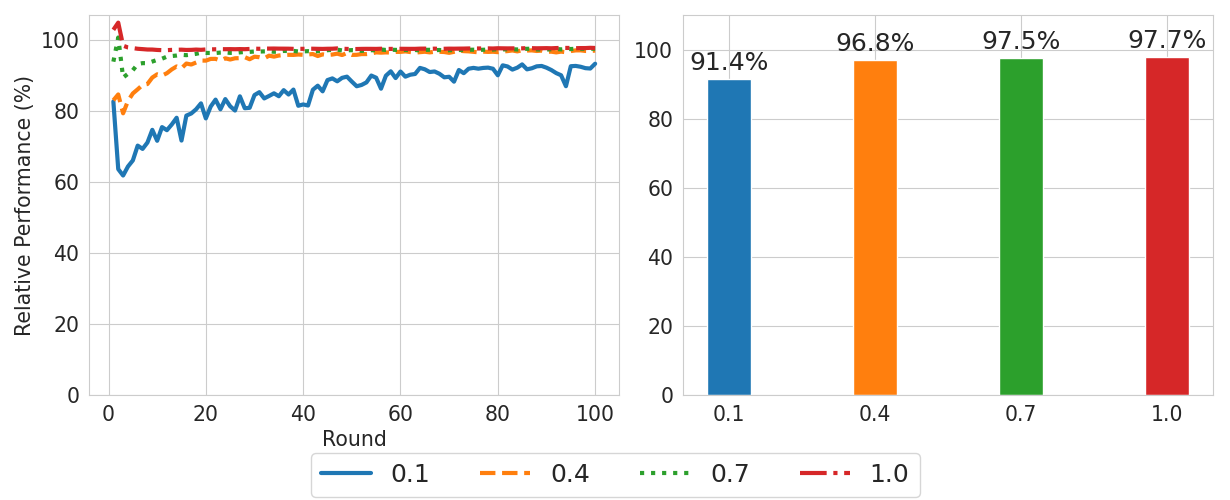}
    \caption{Accuracy trajectories $R(t)$ over 50 rounds (left) and final accuracies $R(50)$ (right) under different participation ratios.}
    \label{fig:app_participate}
\end{figure}

Figure~\ref{fig:app_participate} shows results under Dirichlet($p=0.1, \alpha=0.001$) with varying client participation ratios (0.1, 0.4, 0.7, 1.0). 
Lower participation ratios lead to slower convergence, indicating that the two-stage method does not fully overcome participation-induced variance.

\section{Baseline Settings}
\label{app:baseline}

Our baseline experiments are conducted under configurations that are **comparable to or larger in scale** than those used in recent FL benchmarks.
Details of the shared dataset and client setup are provided in Appendix~\ref{app:baseline_dataset}, and the hyperparameter configurations of each baseline method are described in Appendix~\ref{app:baseline_parameters}.

\subsection{Dataset Settings}
\label{app:baseline_dataset}

Our experiments are conducted on CIFAR-100 and TinyImageNet with 100 clients trained for 50 rounds, a scale comparable to or larger than recent FL benchmarks (see Table~\ref{tab:fl-benchmark-survey}).
\begin{table}[t]
\centering
\footnotesize
\setlength{\tabcolsep}{4pt}
\resizebox{0.7\linewidth}{!}{%
\begin{tabular}{lcc}
\toprule
\textbf{Work} & \textbf{Dataset} & \textbf{\# Clients} \\
\midrule
FedProx~\citep{li2020federated} & MNIST, FEMNIST, Sent140, Shakespeare & 10\\
SCAFFOLD~\citep{karimireddy2020scaffold} & EMNIST & 20\\
FedLTF~\citep{zhan2025fedltf} & CIFAR-10/100, MNIST/FMNIST & 20\\
PFPT~\citep{weng2024probabilistic} & CIFAR-10/100, TinyImageNet & 10\\
FFT-MoE~\citep{hu2025fft} & AgNews, CIFAR-10 & 4, 10\\
\midrule
\textbf{Our setup} & CIFAR-100, TinyImageNet & \textbf{100}\\
\bottomrule
\end{tabular}
}
\caption{Survey of FL benchmarks in recent works. 
Our setup adopts 100 clients on CIFAR-100 and TinyImageNet, 
which is comparable to or larger than prior scales.}
\label{tab:fl-benchmark-survey}
\end{table}

\subsection{Baseline Parameters}
\label{app:baseline_parameters}

Baseline methods are reproduced using their original model architectures and training protocols as specified in their respective papers. 
We preserve the encoders used in the original implementations (e.g., ViT-B/32, ViT-B/16), ensuring faithful reproduction of the reported behaviors. 
Detailed configurations are summarized in Table~\ref{tab:baseline-params}.

\begin{table}[t]
\centering
\footnotesize
\setlength{\tabcolsep}{6pt}
\begin{tabular}{l|c|c}
\toprule
 & PFPT & FFT-MoE \\
\midrule
Batch size & 16 & 128 \\
Encoder & ViT-B/32 & ViT-B/16 \\
Optimizer & Adam $(\beta=(0.9,0.98), \epsilon=1e^{-6})$ & Adam (weight decay=$1e^{-2}$) \\
Learning rate & $1e^{-4}$ & $3e^{-4}$ \\
Local epochs & 5 & 1 \\
Total rounds & 50 & 50 \\
Active client ratio & 0.1 (10/100) & 1.0 (full) \\
\bottomrule
\end{tabular}
\caption{Detailed training configurations of the baseline methods. 
PFPT: number of tokens = 10. 
FFT-MoE: num\_experts = 8, rank\_per\_expert = 2, top-k = 1, auxiliary loss $\lambda = 10^{-5}$.}
\label{tab:baseline-params}
\end{table}

\end{document}